\documentclass[sigconf]{acmart}
\AtBeginDocument{%
  \providecommand\BibTeX{{%
    \normalfont B\kern-0.5em{\scshape i\kern-0.25em b}\kern-0.8em\TeX}}}

\copyrightyear{2021}
\acmYear{2021}
\setcopyright{acmlicensed}\acmConference[SIGCSE '21]{Proceedings of the 52nd ACM Technical Symposium on Computer Science Education}{March 13--20, 2021}{Virtual Event, USA}
\acmBooktitle{Proceedings of the 52nd ACM Technical Symposium on Computer Science Education (SIGCSE '21), March 13--20, 2021, Virtual Event, USA}
\acmPrice{15.00}
\acmDOI{10.1145/3408877.3432439}
\acmISBN{978-1-4503-8062-1/21/03}

\usepackage{subfiles}
\usepackage{caption}
\usepackage{subcaption}
\usepackage[flushleft]{threeparttable}

\usepackage{relsize}
\usepackage{xcolor}

\begin{document}
\fancyhead{}
\title{Early Performance Prediction using Interpretable Patterns in Programming Process Data}



\author{Ge Gao}
\affiliation{%
  \institution{North Carolina State University}
  \city{Raleigh}
  \country{USA}}
\email{ggao5@ncsu.edu}
\author{Samiha Marwan}
\affiliation{%
  \institution{North Carolina State University}
  \city{Raleigh}
  \country{USA}}
\email{samarwan@ncsu.edu}
\author{Thomas W. Price}
\affiliation{%
  \institution{North Carolina State University}
  \city{Raleigh}
  \country{USA}}
\email{twprice@ncsu.edu}



\begin{abstract}
Instructors have limited time and resources to help struggling students, and these resources should be directed to the students who most need them. To address this, researchers have constructed models that can predict students' final course performance early in a semester. However, many predictive models are limited to static and generic student features (e.g. demographics, GPA), rather than computing-specific evidence that assesses a student's progress in class. Many programming environments now capture complete time-stamped records of students' actions during programming. In this work, we leverage this rich, fine-grained log data to build a model to predict student course outcomes. From the log data, we extract patterns of behaviors that are predictive of students' success using an approach called differential sequence mining. We evaluate our approach on a dataset from 106 students in a block-based, introductory programming course. The patterns extracted from our approach can predict final programming performance with 79\% accuracy using only the first programming assignment, outperforming two baseline methods. In addition, we show that the patterns are interpretable and correspond to concrete, effective -- and ineffective -- novice programming behaviors. We also discuss these patterns and their implications for classroom instruction.

\end{abstract}







\maketitle
\vspace{-0.05 in}
\section{Introduction}

Many students struggle during introductory programming courses, failing classes \cite{bennedsen2019failure, watson2014failure} or dropping out of CS programs \cite{chen2013stem}. One-on-one instructor guidance is one of the most effective ways to help struggling students \cite{bloom19842}, but instructors have limited time to spend with individuals. What if an instructor could know after a student's first programming assignment whether the student will be a low performer in the course? This would allow an instructor to focus their finite resources on helping the students who stand to benefit most from them. It would also allow researchers to personalize  automated help (e.g. hints and feedback \cite{price2017hint, rivers2013automatic}), to offer higher levels of scaffolding only to students who need it.

Prior work has constructed machine learning approaches to predict students' performance in a given course based on \textit{static} and \textit{general} factors, such as grades in their previous courses, majors, and demographic factors \cite{elgamal2013educational, hu2019reliable}. These models have achieved some predictive success. However, few models use factors specific to Computer Science, such as students' programming behavior. Additionally, by focusing on static features (e.g. GPA, demographics), these models might suggest that a student's potential to succeed is fixed, rather than a dynamic product of their in-class effort. There is also a need for more \textit{interpretable} predictive models, which can tell us not only \textit{whether} a student will succeed, but also offer insight into \textit{why}, with implications for pedagogy.

Many programming environments capture rich log data as students work, including detailed, sequential records of students' interactions with the environment \cite{hosseini2014exploring, piech2015deep, price2017isnap}. Prior work has shown that analysis of this \textit{process data} can lead to more accurate predictions of student performance than static measures (e.g. assignment grades) \cite{blikstein2014, hadwin2007examining, piech2015deep}. Log data also captures the same information that an instructor might observe when watching a student program, suggesting potential for data-mined patterns to be interpretable by instructors. Many existing approaches for predicting students' final performance with process data use expert-defined metrics, for example based on students' patterns of compiler errors \cite{becker2016new,carter2015towards}). However, these metrics require extensive researcher expertise to author manually, and are only weakly predictive of student outcomes \cite{price2020iticse}.

To address these limitations, we propose a novel, \textit{data-driven} approach for predicting student course outcomes based on their \textit{process data} (i.e. interactions with the programming environment). We extract patterns of students’ actions from their log data on \textit{early assignments}, which are predictive of their \textit{final} performance in a course. We evaluate our approach in a block-based programming course, finding that some of these patterns correspond to meaningful novice programming behaviors and present case studies of how they manifested in individual students. Specifically, we investigate two research questions: 

\textbf{RQ1:} How do students' patterns of programming predict their final performance in a course, and how early can we make this prediction? 

\textbf{RQ2:} How do the patterns generated by such an approach inform our understanding of how students struggle and learn to program? 


Our \textbf{contributions} are three-fold: (a) we present a sequence mining method for programming log data, based on effective approaches in other domains; (b) we demonstrate that our data-driven approach can generate patterns that predict students final course performance with $\sim$79\% accuracy, using log data from just the first assignment; (c) we show how patterns extracted by our approach can inform our understanding of students' programming behavior.

\vspace{-0.05 in}
\section{Related Work}

Many approaches have been proposed for making early predictions of student outcomes in CS courses. These approaches generally consist of two steps: 1) extracting relevant features and 2) training a machine learning model to predict outcomes. Here we focus on the \textit{feature extraction} step. These features can be generally categorized into four types:

\textbf{Background Features}: These are static features about a student or their history before entering the course, such as demographics, GPA, and prior courses taken \cite{elgamal2013educational, hu2019reliable}. For example, ElGamal et al. predict students' programming performance based on their demographics such as gender and prior experience \cite{elgamal2013educational}. However, these approaches require that instructors collect this data, which is not always done. Additionally, these features only reflect a student's background, and not their effort in the course, and their predictions may therefore promote an unproductive \textit{fixed mindset}, rather than a \textit{growth mindset} for instructors who use them \cite{murphy2008dangers}.

\textbf{Grade Features}: These features use students' grades on prior assignments \textit{within a given course} to predict later course outcomes \cite{castro2017evaluating, romero2008data}. For example, Castro-Wunsch et al. counted passed and failed assignments as predictors of struggling students \cite{castro2017evaluating}. These features do capture students' in-course performance and effort; however, student grades only describe a students' final product, and lose relevant information about \textit{how} the student created it.

\textbf{Expert-authored features derived from Process Data}: These features are \textit{defined by experts} and extracted from students' \textit{process data} as they work, e.g. on programming assignments \cite{ahadi2015exploring, castro2017evaluating, emerson2019predicting} or clicker questions \cite{liao2019robust, porter2014predicting}. They might be simple counts of student actions (e.g. steps taken \cite{ahadi2015exploring} or submissions \cite{emerson2019predicting}), or more complex metrics derived from student action sequences. For example, error metrics, such as the Error Quotient \cite{jadud2006methods}, NPSM \cite{carter2015towards, carter2017using} and others \cite{watson2013predicting, becker2016new} quantify the extent to which students struggle with error metrics based on a sequence of their compilation and other behaviors (e.g. create break points \cite{carter2015towards}). However, these features require careful expert authoring, based on observations of student behavior, and as such they may not generalize to new datasets, and can have low predictive accuracy \cite{price2020iticse}.

\textbf{Data-mined features extracted from Process Data}: These features are extracted \textit{automatically} (or semi-automatically) from students' programming process data \cite{wangcomparing}, \textit{discovering} behaviors that might be predictive of student success, beyond what experts have identified. For example, Blikstein, Piech, et al. \cite{piech2012modeling, blikstein2014} used learning analytics approaches to cluster students' problem-solving trajectories on programming assignments and found these clusters to be predictive of students' final grades, moreso than midterm grades. Mao et al. used temporal pattern mining approaches to extract features from sequences of programming snapshots \cite{mao2020time}. These models showed strong predictive success, though they predicted performance on a \textit{single assignment}, rather than across the semester, as in this work. From a theoretical perspective, Grover et al.'s hypothesis-driven framework for learning analytics \cite{grover2017framework} argues that students' fine-grained interactions can serve as meaningful evidence of specific competencies, arguing that ``students’ actions need to be detected and measured as students work.'' They demonstrate how patterns mined from student process data can also be used to inform our understanding of how students express those competencies, though they stop short of using these mined patterns in their assessment models.

In this work, we compare data-mined process data features to a baseline model that uses grade and expert-authored features to determine whether data-driven features improve our ability to predict student outcomes.

\vspace{-0.05 in}
\section{Method}



We present $\chi^2$ - differential sequence mining (CDSM), a data-driven feature engineering algorithm, designed to identify sequences of events (i.e. patterns) in students' log data that differentiate two groups of students: high-performing (\textbf{HP}) and low-performing (\textbf{LP}) students. Specifically, the method identifies two types of patterns: 1) patterns that occur for \textit{more students} in one group than another (e.g. 30\% of LP students vs. 60\% of HP students), and 2) patterns that occur for students in both groups, but appear \textit{more times} for students in one group than another (e.g. an average 5 times per assignment for HP students vs. 1 time per assignment for LP students). The presence of these patterns can then be used as features in a predictive model. 

CDSM builds on the original differential sequence mining (DSM) approach of Kinnebrew et al. \cite{kinnebrew2013contextualized}. We extend this approach in 3 ways. First, we define an \textit{event categorization} method for extracting discrete, meaningful events from programming process data (Section~\ref{subsec: event_categorization}). We then extend DSM to detect less frequent, but still discriminating patterns using a $\chi^2$ test (Section~\ref{subsec:pattern_mining}). Last, we use feature discretization to improve the performance of the extracted features in a model (Section~\ref{subsec:discretization}).

\subsection{Event Categorization}
\label{subsec: event_categorization}

For each student in class, their log data can be represented as a sequence of discrete events that occurred during programming assignments, along with a label (e.g. high/low performing). The sequence should capture all student interactions with the programming environment (e.g. editing code), as well as other interface actions (e.g. menu clicks). The first challenge for the algorithm is to represent these events with the right level of granularity. For example, if events are too specific (e.g. treating every keystroke as a unique event) it will be difficult to find shared patterns across students, but if they are too general (e.g. one event for all code edits), patterns may not be meaningful.

To address this, we leverage the event categorization scheme of the ProgSnap2 format for programming log data \cite{price2020iticse} to make our approach more generalizable across datasets, as ProgSnap2 is designed to represent students' interactions within a variety of programming environments (e.g. block- and text-based) and educational settings. This allows us to build on established event definitions and makes it straightforward to apply our approach to a growing number of programming datasets using ProgSnap2 format. 
Specifically, we used 5 types of events\footnote{We used a subset of ProgSnap2 events relevant to our block-based dataset, but additional events, such as ``Compile'' could be included for other datasets.}: 1) \textbf{EDIT} (File.Edit): The student edits their program. ProgSnap2 also defines several subcategories of EDIT events, and we treat these as distinct events: EDIT-INS: code is inserted into the workspace, EDIT-DEL: code is deleted from the workspace, and EDIT-PST: code is pasted or duplicated into the workspace; 2) \textbf{RUN} (Run.Program): The student runs their program; 3) \textbf{FILE} (File.*): The student performs a file-related action, such as closing, saving, and deleting a file. We collapse these various ProgSnap2 events into a single FILE event type; 4) \textbf{CHAN} (X-ChangeBlockCategory): The student selects a different block category (events prefixed with `X-' are specific to our dataset, described in Section~\ref{subsec:population}); 5) \textbf{VAR} (X-AddVariable): The student creates a new variable.

    

\vspace{-0.1 in}
\subsection{Pattern Mining}
\label{subsec:pattern_mining}

\begin{figure}[t]
\includegraphics[scale=0.17]{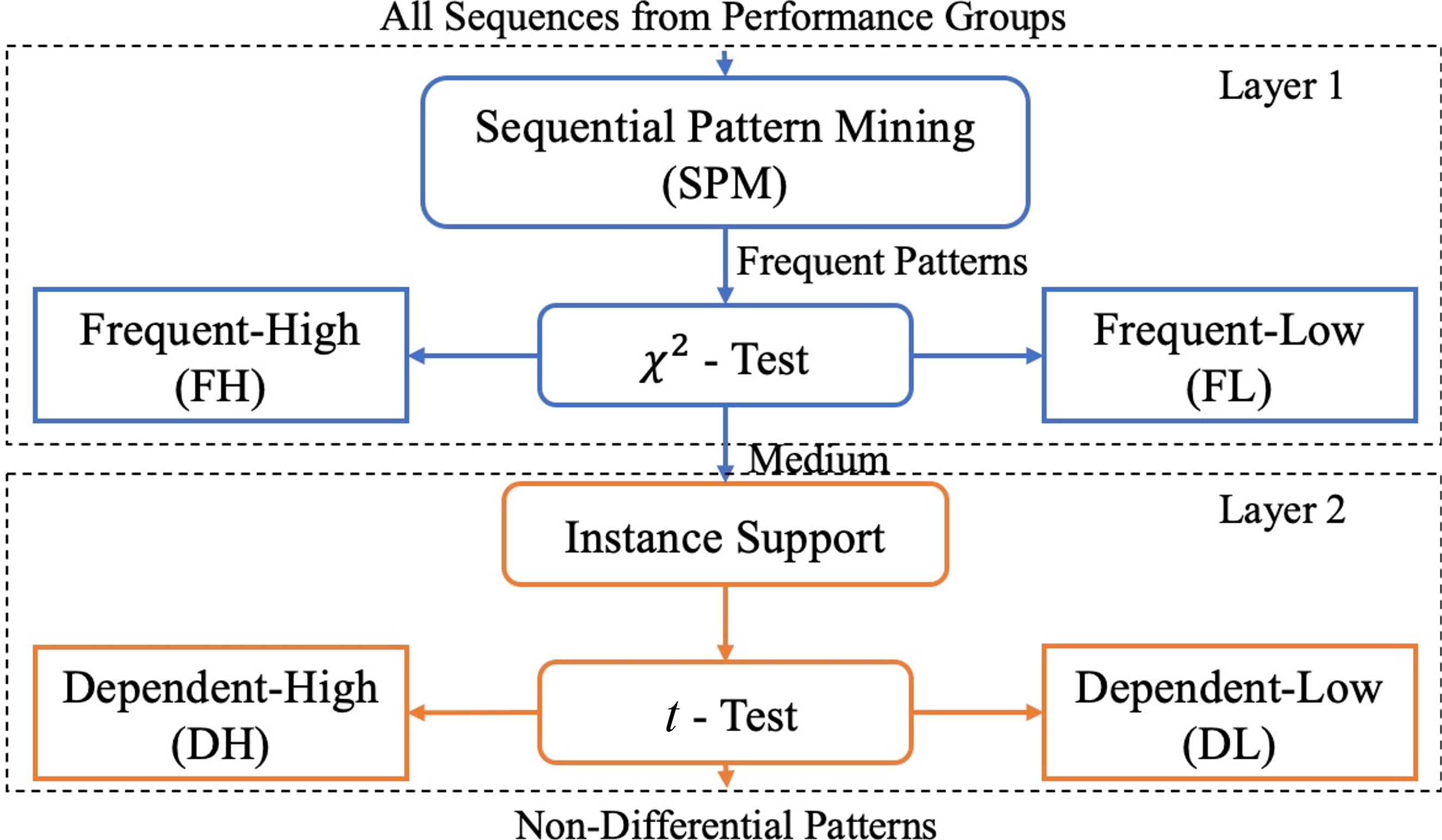}
\centering
\caption{Schematic of the Pattern Mining Algorithm. 
}
\label{fig:cdsm}
\vspace{-0.2 in}
\end{figure}

After categorizing all events in the trace data, all the actions taken by students in each programming assignment can be formulated as a sequence of events. To simplify the representation of sequences, consecutive actions of the same type are collapsed into a single action in this paper (e.g. EDIT RUN is considered instead of EDIT EDIT RUN). This is useful because it collapses similar patterns into one representation, making it easier to find common behaviors among students. However, this also means we may lose some potentially relevant details in the patterns. This choice is designed to strike a balance between specificity and generality. Then the pattern mining step is devised to identify patterns, from these sequences, which can differentiate programming behaviors between two students groups (i.e. HP and LP )regarding their final performance.\footnote{Certain criteria (e.g. course final grade) can be used to divide students into these two groups. } 

To achieve this, we propose a two-layer pattern mining algorithm as depicted in Figure~\ref{fig:cdsm}. In Layer 1, Sequential Pattern Mining~\cite{agrawal1995mining} is first applied to find frequent patterns (\textbf{FP}s) among all the sequences within each performance group, respectively. Then a $\chi^2$-test is performed to identify two types of patterns -- \textit{frequently} high performance (\textbf{FH}) or low performance (\textbf{FL}) , which contain patterns that \textit{frequently} occurred among HP and LP students respectively. The patterns do not show statistical significance in this layer are assigned to the `Medium' group and transmitted to Layer 2. Specifically, we extract FPs according to 3 factors -- minimum percentile support, maximum gaps and maximum length. Minimum percentile support is the minimum portion of sequences that contains a pattern. Maximum gaps is the maximum number of actions can be skipped when constructing patterns within a sequence. Maximum length refers to the maximum number of actions required to construct a pattern within a sequence. Then, all the patterns whose all three factors meet certain thresholds are selected to be FPs. Next, for each FP, we count the frequency of occurrence (FoC) in the HP and LP groups respectively. $\chi2$-test is employed to investigate whether the FoC of a FP \textit{significantly} correspondeds to any performance groups (i.e. HP or LP). If so, the FP is associated categorized as FH or FL.               

In Layer 2, the instance-support~\cite{lo2008efficient} of each FP in the Medium group is calculated, followed by a $t$-test to label it as a dependent-high (\textbf{DH}) or dependent-low (\textbf{DL}) pattern. Specifically, for each pattern, its instance-supports are calculated across all sequences in HP and LP respectively, which then results in two sets of instance-support (i.e. one set corresponds to HP and the other set corresponds to LP). Then, a $t$-test is used to determine whether one set is \textit{significantly} different from the other. If so, then it implies that the corresponding FP appears more frequently in one performance group than the other, and it is associated with either DH or DL depending on the mean value of instance-supports within each group (we select the group with higher mean value). If the $t$-test does not show significance, then the corresponding FP is determined as non-differential patterns and discarded.

\vspace{-0.1 in}
\subsection{Feature Discretization}
\label{subsec:discretization}

As prediction techniques usually require numeric inputs, the patterns we obtained from the pattern mining procedure need to be cast as numerical features. To achieve this, we design a tabular feature representation method. Specifically, we store the FoC of all FPs in the 4 categories in a feature table where each column corresponds to a FP and each row corresponds to a student. 
We use discretization in this work since it has low computational complexity and has generally shown a good performance on classification tasks \cite{frank1999making}. Specifically, for the columns that are associated with FH or FL, all the containing FoCs are discretized into two equal-frequency bins. The cells that are smaller than the median are assigned with 0's, while the others are assigned with 1's. Similarly, for columns belonging to DH and DL groups, all the FoCs are discretized into three equal-frequency bins with values 0, 1, and 2 using 1/3 and 2/3 quantiles. We split them up into 3 bins because the FPs in DH and DL are more common to appear in both student performing groups (i.e. FH and FL) simultaneously.




\vspace{-0.05 in}
\section{Experiment}

\subsection{Population and Data}
\label{subsec:population}

We use data from an introduction to computing course for non-majors at a large, public university in the Eastern United States. The instructors were not part of the research team. The data was collected across two semesters, Spring 2017 and Fall 2017, with a total of 106 students. In both semesters, students worked on the same 5 programming exercises (A1-A5) in a block-based programming environment called Snap. 
We did not have access to students' demographic information or prior experience, but the course was designed for novices. The programming part of the course was taught in labs led by undergraduate TAs. As students worked, the programming environment automatically logged each action a student took (e.g. adding a code block, running code), including complete snapshots of students' code, allowing researchers to recreate a students' programming process.

Rather than using in-class grades which are made by undergraduate TAs (which vary year-to-year), we regraded the exercises with a similar set of rubrics, using the final snapshot of each student’s program, based on the exercise's objectives. We treat the average of students’ grades on these exercises as a measure of their programming performance over the whole course, with the grade of each exercise scaled to (0, 1). We used a median split to divide the students into 54 high performers and 52 low performers, based on a median grade of 0.84. As noted in prior work, a median split is appropriate for evaluating the performance of the classification models as well as identifying the patterns that can significantly differentiate high and low performers \cite{ames1988achievement, ghantasala2013median}. This aligns with our goal of identifying students who will perform worse than others and might benefit from prioritized help (not only students `at-risk' of failing). Our log data is in a well-defined data sharing format, ProgSnap2 \cite{price2020iticse}, and we use it to extract patterns by CDSM. In this study, our data consists of $\sim$530 sequences and $\sim$500 events per sequence.

\vspace{-0.1 in}
\subsection{RQ1: Performance Prediction}
\label{subsec:prediction}

Since RQ1 asks how early we can predict students' grades, we performed 5 separate prediction trials (M1-M5). For a given trial, $M_i$, we only used data from assignments $A_1$ through $A_i$. All trials predicted students' final performance. We extracted patterns from each assignment individually, as completing each assignment requires different programming knowledge, and students’ behavior can change as the course progresses. 
We adopted a \textit{feature stacking} approach, which combines features across assignments to construct a meta feature table. 
We used an Ada-boost classifier and validated it with a modified hold-out 10-fold cross validation, where 8 sets are used for training, 1 set for testing while the remaining one is discarded. 

Our CDSM approach largely uses generic events which exist in any IDE. We wanted to explore how adding IDE-specific context might improve the CDSM approach. We therefore evaluated 2 variants of the Event Categorization approach detailed in Section \ref{subsec: event_categorization}: the \textit{general} approach, which only captures generic event information (as described earlier), and the \textit{contextual} approach, which also captures a small amount of Snap-specific contextual information.  Specifically, contextual events append a suffix to each action with the name of currently opened block category in Snap (describing what type of blocks a student can add). For example, when an EDIT-INS-pen action is recorded, it means that a student is programming in Snap with \textit{Pen} category opened. Consequently, two explanation levels -- \textit{general} and \textit{contextual} -- of event categorization techniques can be combined with the CDSM to produce patterns with different levels of explainability, which are denoted as \textbf{CDSM-G} and \textbf{CDSM-C} respectively.


We employ two baselines: 1) \textit{Majority}: using the majority label of performers as predicted label. 
2) \textit{Expert Rule}: using students' grades on prior exercises (e.g. grades of A1 and A2 for trial M3), as well as 5 expert-authored process data features: \textit{Block Deletions} (number of blocks deleted), \textit{Block Moves} (number of blocks moved), \textit{Code Runs} (number of program run events), \textit{Time} (the total minutes spent on programming exercises), and \textit{Meaningful Nodes} (the total number of loops, conditionals, and custom blocks made by students in their code). Each of these features has been found to correlate with student performance in previous research (e.g. \cite{carter2015towards, emerson2019predicting, hsu2003hybrid, leinonen2017comparison}). Note that Snap does not have compile events, so we could not use compilation error metrics as a baseline. We also employed an Ada-boost classifier with the features from the \textit{Expert Rule} baseline, using the same procedure to tune hyperparameters as our approach.    


\vspace{-0.1 in}
\subsection{RQ2: Pattern Interpretation}

We investigated a meaningful subset of the patterns generated by the CDSM approach to evaluate whether they corresponded to meaningful student behaviors. 
To investigate these patterns, we recorded all patterns extracted from CDSM with their corresponding assignment name and performance group. Then for each pattern, we calculated: 1) portions (i.e. $Perc\_High$, $Perc\_Low$) of \textit{all} students who have the pattern among high and low performers respectively; 2) the difference (\textit{Diff}) between $Perc\_High$ and $Perc\_Low$; 3) the Odds Ratio (OR) which shows how much more/less likely students who do the pattern are to be low performers. We sorted the patterns by their \textit{Diff} in descending order. Statistically, the higher \textit{Diff} implies the more likely a pattern happens in a specific group. We selected a set of the top $15\%$ of patterns from both performance groups. This helped us identify the most frequent patterns and avoid missing important patterns. For each pattern in the set, 3 researchers replayed the logged actions of individual students who performed the pattern to explore how and in what context the pattern happened during students’ programming process. Each researcher summarized their understanding of the pattern independently, then they discussed their interpretations with 2 criteria: 1) what are students doing within the pattern? 2) how might the pattern be indicative of student performance? If all researchers noted that a pattern reflected was indicative of high- or low-performing students, this suggested that the pattern may be meaningful and interpretable.

\vspace{-0.05 in}
\section{Results and Discussion}
\begin{table}[t]
\begin{threeparttable}
\small
\caption{Accuracy, Precision, and Recall of the CDSM-based Prediction Models and Baselines}
\label{table:predRes}
\begin{tabular}{p{0.2cm}p{0.2cm}p{0.2cm}p{0.2cm}p{0.2cm}p{0.2cm}p{0.2cm}p{0.2cm}p{0.2cm}p{0.2cm}p{0.2cm}p{0.2cm}p{0.2cm}}
\hline
                        & \multicolumn{4}{c}{Accuracy}                   & \multicolumn{4}{c}{Precision}                  & \multicolumn{4}{c}{Recall} \\ \hline
\multicolumn{1}{c|}{} & b.1 & b.2 & G    & \multicolumn{1}{c|}{C}    & b.1 & b.2 & G    & \multicolumn{1}{c|}{C}    & b.1  & b.2 & G    & C    \\ 
\multicolumn{1}{l|}{M1}   & .509 & .477 & .792 & \multicolumn{1}{l|}{.792} & .509 & .509 & .833 & \multicolumn{1}{l|}{.808} & 1     & .558 & .769 & .808 \\
\multicolumn{1}{l|}{M2}   & .509 & .617 & .773 & \multicolumn{1}{l|}{.753} & .509 & .617 & .759 & \multicolumn{1}{l|}{.741} & 1     & .698 & .830 & .811 \\
\multicolumn{1}{l|}{M3}   & .509 & .677 & .850  & \multicolumn{1}{l|}{.774} & .509 & .684 & .852 & \multicolumn{1}{l|}{.778} & 1     & .722 & .852 & .778 \\
\multicolumn{1}{l|}{M4}   & .509 & .728 & .819 & \multicolumn{1}{l|}{.803} & .509 & .732 & .797 & \multicolumn{1}{l|}{.780} & 1     & .759 & .870 & .852 \\
\multicolumn{1}{l|}{M5}   & .509 & .778 & .803 & \multicolumn{1}{l|}{.784} & .509 & .763 & .770 & \multicolumn{1}{l|}{.746} & 1     & .833 & .870 & .870 \\ \hline

\hline
\end{tabular}
\begin{tablenotes}
      \small
      \item b.1: Majority; b.2: Expert Rule; G: CDSM-G; X: CDSM-C.
    \end{tablenotes}
\label{tab:accuracy}
\end{threeparttable}
\vspace{-0.2in}
\end{table}

\begin{table*}[t]
\begin{threeparttable}
\caption{Pattern Examples and Corresponding Statistics}
\label{table:pattern}
\begin{tabular}{p{0.8cm}p{0.8cm}p{9cm}p{1cm}p{1cm}p{0.6cm}p{1.4cm}}
\hline
Index & Ex. & \multicolumn{1}{c}{Patterns}                                                        & PercHigh & PercLow & Diff. & OR \\ \hline
HG1  & A2      & EDIT-INS CHAN EDIT-INS EDIT CHAN EDIT-INS                                           & 52\%     & 31\%    & 21\%  & 2.41       \\
LG1  & A1      & EDIT-INS EDIT RUN EDIT-INS EDIT RUN                                                 & 37\%     & 60\%    & 23\%  & 0.34       \\ 
LG2 & A2 & CHAN RUN EDIT-INS & 22\% & 42\% & 20\% & 0.39 \\
\hline
\end{tabular}
\begin{tablenotes}
      \small
      \item Ex.: the exercise which the pattern is extracted from; PercHigh/Low: the percentage of high/low-performers who have the pattern; Diff.: difference between PercHigh and PercLow; OR: odds ratio.
    \end{tablenotes}
\end{threeparttable}
\vspace{-0.15 in}
\end{table*}


\subsection{RQ1: Performance Prediction}

\begin{figure}[t!]
\includegraphics[scale=0.33]{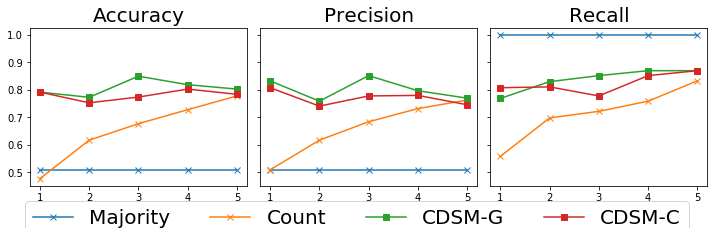}
\centering
\caption{Visualization of Prediction Results by Adaboost. X-Axis: Trial Number; Y-Axis: Prediction Results.
}
\label{fig:pred-result}
\vspace{-0.2 in}
\end{figure}

The prediction results from Adaboost classification using patterns generated from our approach and 2 baselines are shown in Table~\ref{table:predRes} and visualized in Figure~\ref{fig:pred-result}. Our approach (CDSM-G and -C) performs better with higher accuracy and precision than both baselines among all trials across exercises, except precision on the 5th trial (M5) of CDSM-C. Our approach reaches to 0.792 accuracy within the first exercise. Specifically, using data from only the first exercise, our approach can identify more than $ 80\%$ of students who will end up as low performers, while miss-identifying less than $20\%$ of high-performers as low-performers (CDSM-G). By contrast, the Count baseline (which included students' actual grades on prior assignments) only reached that accuracy on the last trial, with all available data. We also find that CDSM-C has less accuracy and precision than CDSM-G across exercises, which indicates that contextual information did not improve the prediction.

\vspace{-0.1 in}
\subsection{RQ2: Pattern Interpretation}

From the researchers' examination of patterns, we note that: 1) Not all of the patterns were very informative. For example, the ``RUN EDIT-ADD'' pattern was more frequent with low performers, but this only indicates that they are more likely to run their code followed by adding blocks, which is a generic behavior to most students. 2) Some of the patterns are highly related (e.g., EDIT-INS CHAN is a subset of EDIT-INS CHAN CHAN) and convey similar information to researchers. We focus on only one pattern from each cluster of similar patterns. Our analysis discovered $\sim15\%$ distinct, meaningful patterns among the 288 we inspected, from more than 1000 total patterns. Here we focus on case studies of 3 examples (shown in Table \ref{table:pattern}), where all researchers agreed on their interpretability and meaningfulness. 

\begin{figure} [t!]
     \centering
     \begin{subfigure}[b]{0.23\textwidth}
         \centering
         \includegraphics[width=\textwidth, height=3cm]{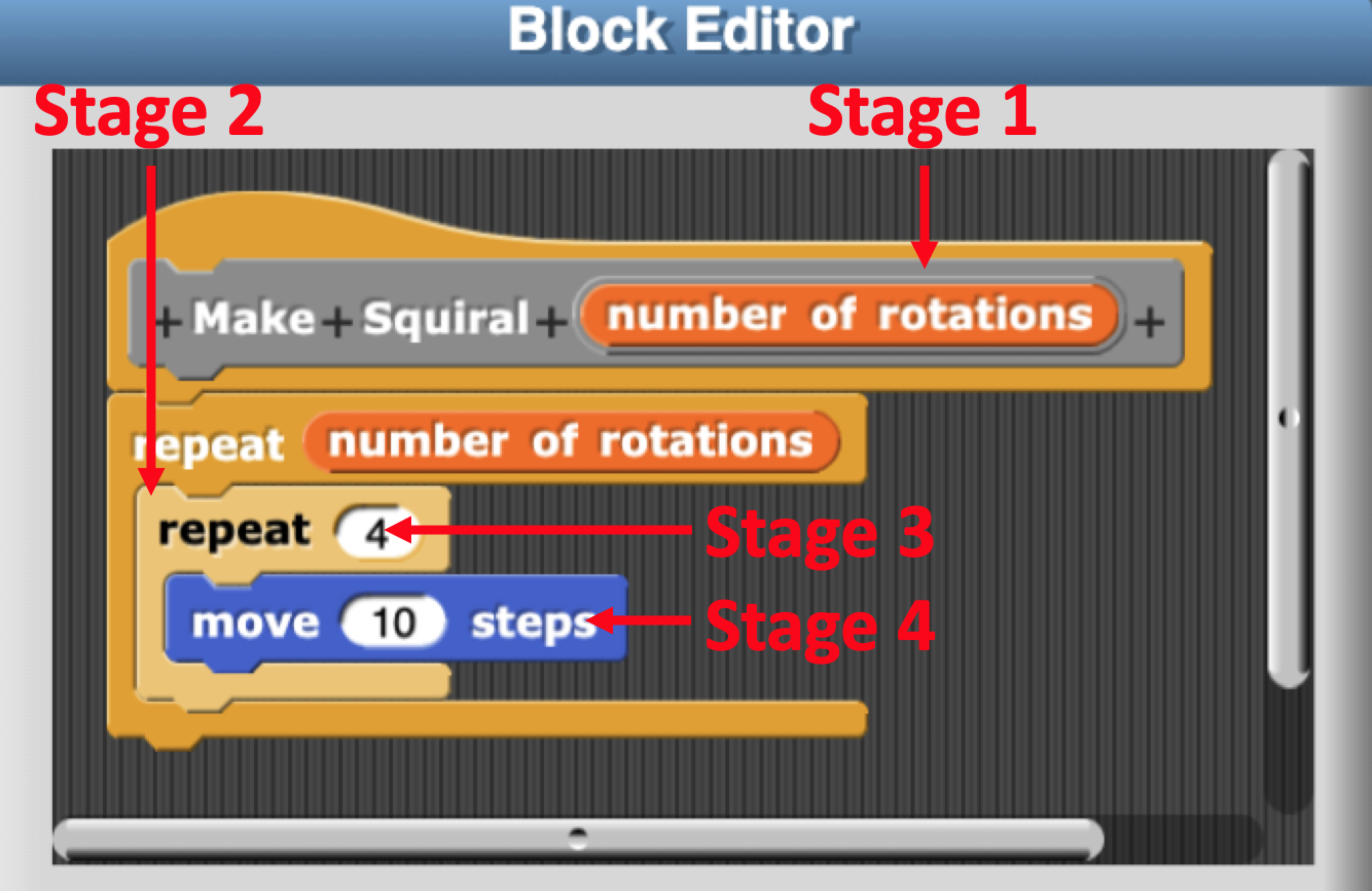}
         \caption{HG1 of Student A. 
         }
         \label{fig:studentA}
     \end{subfigure}
     \begin{subfigure}[b]{0.23\textwidth}
         \centering
         \includegraphics[width=\textwidth, height=3cm]{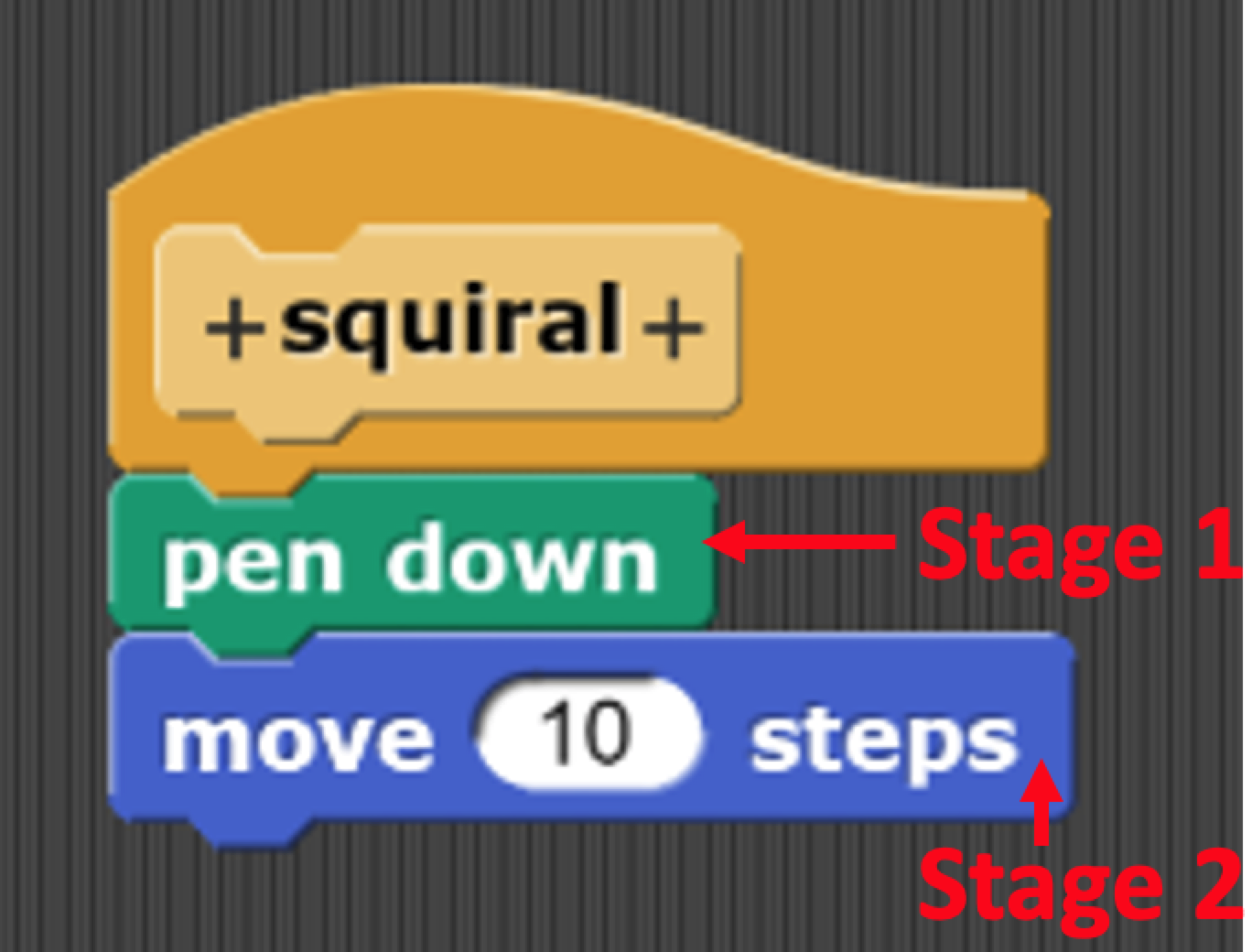}
         \caption{LG2 of Student C. 
         }
     \end{subfigure}
        
     \begin{subfigure}[b]{0.46\textwidth}
         \centering
         \includegraphics[width=\textwidth, height=2.8cm]{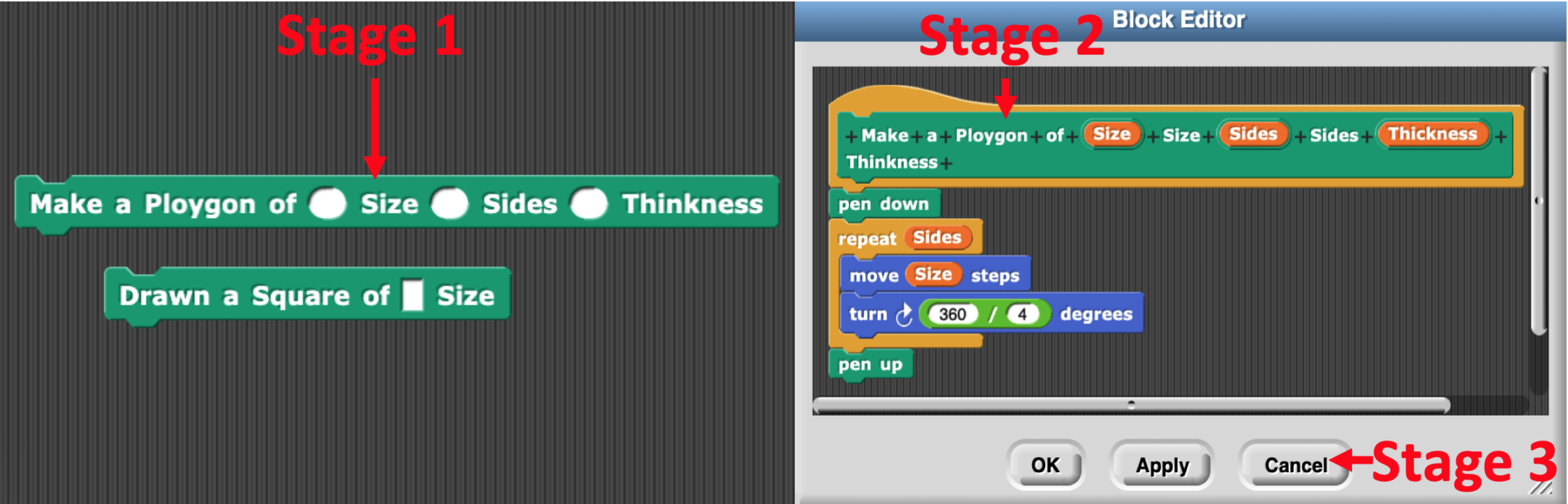}
         \caption{LG1 of Student B. 
         }
     \end{subfigure}
    \caption{Code Construct Examples with Selected Patterns}
    \label{fig:examples}
\vspace{-0.25 in}        
\end{figure}

\subsubsection{Case Study 1: Abstraction}
The HG1 pattern, EDIT-INS CHAN EDIT-INS EDIT CHAN EDIT-INS, was detected for assignment A2 as more common among high performers. In this pattern, students create a custom block (i.e. a procedure), possibly with parameters, then add blocks from two different categories. The reason why the first EDIT-INS specifically indicates creating a new custom block (procedure), as opposed to adding a new \textit{code block}, is that it is not followed by an EDIT, which always occurs when students ``drop'' a new code block into their workspace (i.e. stage). 
HG1 also suggests that students are creating a custom block before they create the code that will go inside of it, as students who move existing code into a newly created custom block would not have the second and third EDIT-INS events. While this pattern may seem overly-specific to one programming environment and its logging approach, CDSM's data-driven approach allows it discover similar niche patterns from other log data too.

Student A is a high performer who shows pattern HG1 in A2. 
Figure \ref{fig:examples} shows the code construct that occurred during pattern HG1 for Student A over 10 seconds. It mainly consists of 4 stages of code changes: 1) EDIT-INS: Create a custom block and a variable named ``number of rotations''; 2) CHAN EDIT-INS: Change to the \textit{Control} category and add a ``repeat'' block with default parameter 10 into stage; 3) EDIT: Configure the ``repeat'' block by updating the parameter to 4; 4) CHAN EDIT-INS: Change to \textit{Motion} category and add a ``move'' block. In this study, the frequency that this behavior happens among high performers is much higher than low performers ($52\%$ vs $31\%$). This could indicate that students with HG1 understood the importance of custom blocks (i.e procedures), and seemed to have a plan for what to do in that block by adding two new blocks right after.

Students who demonstrated HG1 started by defining a procedure, before adding code, suggesting a more-top down design of their program, or at least more familiarity with how to create procedures. Prior studies related such observations to students' abstraction behavior \cite{ye1996expert, wing2006computational}, as procedures allow programmers decompose problems and hide unneeded implementation details. 

\subsubsection{Case Study 2: Debugging}

The LG1 pattern, EDIT-INS EDIT RUN EDIT-INS EDIT RUN, was detected for assignment A1 as more common among low performers than high performers (60\% vs 37\%). In this pattern, students repeatedly add a new piece of code, then edit their code and run it. Recall that a given event (e.g. EDIT) can represent multiple instances of a type of event occurring in a row, so the events in this sequence may represent a duration of editing or running. Student B is a low performer who has the pattern LG1 in the A1 exercise. Figure \ref{fig:examples} shows the code construct captured among student B’s changes on their code within 14 seconds when they are doing LG1. It mainly contains 3 stages: 1) EDIT-INS EDIT RUN: Move a block back-and-forth by grabbing one existing custom block ``Make a Polygon of $\#$ Size $\#$ Sides $\#$ Thickness [sic]'' and snapping it back to the stage. The student then runs code twice but nothing is drawn because of missing parameters inside the procedure; 2) EDIT-INS: Create an incorrect custom block for drawing a polygon because of the wrong usage of ``turn'' blocks. 3) EDIT RUN: Cancel the custom block and run code again without anything drawn. 

We can see that student B runs their program after making one block edit, then they run their program twice after another edit. This is an indicator of debugging behavior with running the program after each single edit. It also suggests that student B does not make progress during the time range when LG1 happens. They move their blocks back-and-forth on stage, which is described as uncertainty or hesitation tinkering behavior by Dong et al. \cite{dong2019defining}, and this is a negative feature of tinkering \cite{perkins1986conditions}. When student B gets stuck, they do not appear to stop and spend time to think over about it or seek help from others. Instead, they act unsystematically, which can make the problem worse \cite{perkins1986conditions}. Beyond student B’s behavior, there are many ways that the pattern LG1 occurs, since it is a general pattern that does not specify block categories. However, it always indicates that a student is running their program frequently within a short period as they enact the pattern. These students may be exhibiting debugging behavior, which has been correlated with lower (B-level) performance \cite{carter2017using}.

\subsubsection{Case Study 3: Testing Blocks Before Adding}

The LG2 pattern, CHAN RUN EDIT-INS, was detected in assignment A2 as more common among low performers (42\% vs 22\%). The pattern consists of a student changing block categories in Snap, then running some code, and then adding new code. For this pattern, it is important to know that in Snap, students can not only run their whole program (by clicking a ``green flag''), but run individual blocks by clicking on them. This includes blocks in the ``palette'', where students can select blocks in opened categories to be added into their code. Manual inspection of the CHAN RUN pattern revealed that the vast majority of the time, it occurs when students switch to a new category and clicked a block in the palette to run it. In other words, the student was not running \textit{their own} code, but running a block they \textit{could add} to their code. This behavior suggests students are unsure what the block does, and may have been searching for a block with a specific functionality. After running the block, the students would often add it to their code. Interestingly, running these blocks sometimes had no visible effect: for example, a ``repeat'' or ``declare variables'' block has no output when run alone.

As an example, student C engages in the CHAN RUN pattern twice at the start of their session. The student first changes to the \textit{Pen} category, then clicked on the ``pen down'' block in the palette which has no visible effect (i.e. they could not see its output). After 6 seconds, they add it to their code. The student then changes to the \textit{Motion} category and runs a ``move'' block in the palette, which moves the sprite forward. After 3 seconds, they add the ``move'' block to their code as well. While these ``pen down'' and ``move'' blocks are relatively common, it is important to remember that the student has only used these blocks in one prior exercise (A1), so they are still relatively new. This behavior was almost twice as common among low performers on A2, suggesting a hypothesis that testing a block before adding it to their code is an evidence of uncertainty or struggle. In addition to the most common case, described above, LG2 also occurs when students run their custom block by clicking on it in the block palette, rather than by adding it to a script in their own code to run it.

Similar observations have been made in prior work. For example, Zhi et al. \cite{Zhi2019ICER} suggest that Parsons problems -- where students are given all the elements of a correct code solution and must rearrange them -- may be effective in part because students do not waste time identifying the code elements they need to solve a problem. This may be particularly true in block-based programming environments, where students must search for new blocks across a large palette. Our results support this interpretation, and they also suggest that uncertainty about the blocks' functionalities, evidenced by the CHAN RUN pattern, may exacerbate this challenge for novices. Mitrovic et al. \cite{mitrovic2013effect} suggest that programming environments can address this uncertainty with positive feedback, e.g. informing students that they have selected the correct block and made progress.



\vspace{-0.05 in}
\section{Conclusion and Limitations}
Overall, the \textbf{contributions} of this work are: (a) we adapt a sequence mining method to programming log data; (b) we demonstrate that our data-driven approach can generate patterns that predict students final course performance with accuracy $\sim$79\%, using log data from just the first assignment; (c) we show how patterns extracted by our approach can inform our understanding of students' programming behavior.

There are two primary limitations in this study: 1) The specific patterns we find are not all generalizable, with some specific to Snap. However, while the individual patterns may not generalize to all contexts, the method should generalize to any context where we have event-level data with ProgSnap2 format. The patterns and our model are evaluated by hold-out 10-fold cross validation, which suggests we will achieve similar results on a new dataset from a similar population. Additionally, many of these events are general across both block-based and textual programming environments. 2) The data does not come from the typical CS class (only 5 programming assignments), and we do not use any external measures (e.g. quizzes). Future work is needed to evaluate the approach in more classroom contexts, including more traditional CS1 courses. 




\bibliographystyle{ACM-Reference-Format}
\bibliography{sigproc}


\end{document}